# Is it a click bait? Let's predict using Machine Learning

## BITS ZG628T: Dissertation

by

Sohom Ghosh

2017HT12194

## Dissertation work carried out at

### Times Internet Limited, Noida, India

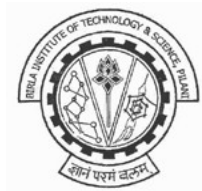

## BIRLA INSTITUTE OF TECHNOLOGY & SCIENCE
## PILANI (RAJASTHAN)

April 2019

# Is it a click bait? Let's predict using Machine Learning

**BITS ZG628T: Dissertation**

by

Sohom Ghosh

2017HT12194

## Dissertation work carried out at

### Times Internet Limited, Noida, India

Submitted in partial fulfillment of M.Tech. Software Systems degree programme

Under the Supervision of

Asif Iquebal Ajazi, Technical Architect
Times Internet Limited, Noida, India

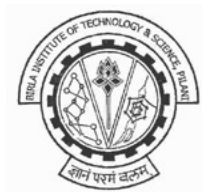

## BIRLA INSTITUTE OF TECHNOLOGY & SCIENCE
## PILANI (RAJASTHAN)

April, 2019

# CERTIFICATE

This is to certify that the Dissertation entitled **'Is it a click bait? Let's predict using Machine Learning'** and submitted by **Sohom Ghosh** having ID-No. **2017HT12194** for the partial fulfillment of the requirements of **M.Tech. Software Systems** degree of BITS, embodies the bonafide work done by him under my supervision.

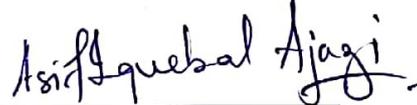

Signature of the Supervisor

Asif Iquebal Ajazi, Technical Architect
Times Internet Limited, Noida, India

Name, Designation & Organization & Location

Place : Noida

Date : 28th March 2019

**Birla Institute of Technology & Science, Pilani**

**WorkIntegrated Learning Programmes Division**

**Second Semester 20182019**

**BITS ZG628T: Dissertation**

**ABSTRACT**

**BITS ID No.** **: 2017HT12194**

**NAME OF THE STUDENT** **: SOHOM GHOSH**

**EMAIL ADDRESS** **: sohom1ghosh@gmail.com**

**STUDENT'S EMPLOYING** **: TIMES INTERNET LTD, NOIDA  201301, INDIA**
**ORGANIZATION & LOCATION**

**SUPERVISOR'S NAME** **: ASIF IQUEBAL AJAZI**

**SUPERVISOR'S EMPLOYING** **: TIMES INTERNET LTD, NOIDA  201301, INDIA**
**ORGANIZATION & LOCATION**

**SUPERVISOR'S EMAIL ADDRESS: asif.iquebal@gmail.com**

**DISSERTATION  TITLE** **: IS IT A CLICKBAIT? LET'S PREDICT USING**
**MACHINE LEARNING**

_________________________________________________________

**ABSTRACT :**  In this era of digitisation, news reader tend to read news online. This is because, online media instantly provides access to a wide variety of content. Thus, people don't have to wait for tomorrow's newspaper to know what's happening today. Along with these virtues, online news have some vices as well. One such vice is presence of social media posts (tweets) relating to news articles whose sole purpose is to draw attention of the users rather than directing them to read the actual content. Such posts are referred to as clickbaits. The objective of this project is to develop a system which would be capable of predicting how likely are the social media posts (tweets) relating to new articles tend to be clickbait.

**Broad Academic Area of Work: MACHINE LEARNING**

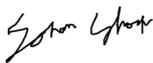

-----------------------------
Signature of the Student
Name: Sohom Ghosh
Date: 14-01-2019
Place: Noida

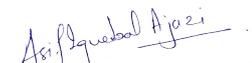

-----------------------------------
Signature of the Supervisor
Name: Asif Iquebal Ajazi
Date: 14-01-2019
Place: Noida

# Abstract

Over the years, the way people reads news is changing. Instead of newspapers in printed form, millennials prefer to read them online. Another reason for the massive popularity of digitised news is unlike traditional newspapers, that they provide instant information about the happenings around the globe. Online news has its' own nuisances. A major chunk of users visits news websites by reading titles / short descriptions of news on social media platforms. 'All that glitters is not gold' - Many times, social media posts of digital marketing executives regarding news articles does not reflect the actual content of news. They are mainly written in a way capable of inciting users to click on them thereby increasing the overall traffic of the news websites. Clickbait content refers to those content which is mainly written for attracting the attention of masses. These alluring contents provoke users to click on links and redirect them to land on a particular web page. The objective of click-bait articles is to increase traffic of a website rather than providing quality content to users. It is of utmost importance to check that the nature of social media posts done by digital marketing executives regarding news articles published is not clickbait. This project aims at detecting social media posts which tend to be click-bait.

# Acknowledgements

Firstly, I would like to express the deepest gratitude to all my mentors & managers - Dr Swati Agarwal, Mr Asif Iquebal Ajazi, Mr Uttam Kumar Pandey and Mr Yashu Kant Gupta. I am grateful to Mr Sanjay Goyal & Mr Vivek Pandey for helping me to be a part of the amazing teams of Times Internet. A special thanks for my colleagues at Times Internet and faculty(ies) of BITS, Pilani for their valuable feedbacks.

| Acronym | Full Form |
|---------|-----------|
| CB | ClickBait |
| CSV | Comma Separated Values |
| JSON | JavaScript Object Notation |
| MB | MegaBytes |
| GB | GigaBytes |
| # | Number |
| % | Percentage |
| MSE | Mean Squared Error |
| ROC | Receiver Operating Characteristic |
| LR | Logistic Regression |
| RF | Random Forest |
| XGB | Extreme Gradient Boosting |
| GBM | Gradient Boosting Machine |
| LGBM | Light GBM |
| AUC | Area under ROC curve |
| ML | Machine Learning |
| DL | Deep Learning |
| CNN | Convoluted Neural Networks |
| LSTM | Long Short Term Memory |
| GRU | Gated Recurrent Unit |
| RNN | Recurrent Neural Networks |
| DNN | Deep Neural Networks |
| ReLU | Rectified Linear Unit |

Table 1 List of Acronyms

**Thesis Publication**

Sohom Ghosh (2019). Identifying Clickbaits using various Machine Learning and Deep Learning Techniques. [14] The $2^{nd}$ International Conference On Information Technology and Digital Applications 2019 [**Accepted For Publication**] [1]

---

[1]This article is included in the thesis chapters.

# Table of contents

# List of figures

# List of tables

# Chapter 1

# Introduction

## 1.1 Context

Recently, the reading habits of news readers are changing. Now they prefer to read news online. This is because the Internet is such a media which provide instant access to a wide variety of content. Thus, they do not have to wait for tomorrow morning's newspaper to know what is happening today. Furthermore, various other features of online media (like the personalization of content, sharing news articles with friends, commenting & discussing these articles in social media) is increasing its' popularity. However, online media has its' own inconvenience. One such inconvenience is the presence of social media posts about news articles whose sole purpose is to increase traffic of news publishers' websites by alluring news readers. The expectations which get set among them while going through these posts relating to what the actual content would be and the actual news content, in reality, does not match. Thus, these kinds of posts make regular news readers feel dissatisfied. The purpose of this project aims at developing a classifier capable of detecting click-bait posts.

## 1.2 Click baits

### 1.2.1 What click-bait articles refer to?

Clickbait content refers to those whose purpose is to attract the attention of people. These contents tend to instigate users to click them. The objective of this kind of articles is to increase the number of visitors to the publisher's website rather than providing users with quality content to read.

### 1.2.2   Why is it essential to identify click-baits?

The reputation of a digital news publisher hugely rests on its' users' opinions. A user tends to re-visit a news portal only when his thirst for latest happenings is quenched properly. If his needs are not catered properly, he may avoid visiting the website again. This reduces the count of daily & monthly active users and impacts business severely. Hence, it is of utmost importance to verify that the nature of social media posts relating to news articles do not tend to be clickbait.

## 1.3   Problem Definition

Given text content, title, description and keywords/tags of a news article, captions of images present in it and a social media post (tweets) relating to it, a system needs to be developed which is capable of predicting how likely is this post (tweets) tend to be clickbait.

# Chapter 2

# Literature Review

In this section, works related to detection of click baits which have already been done are explored.

Elyashar et al. [12] have created some innovative features like counting the number of formal English words, extracting text from images using optical character recognition (OCR). Moreover, they have also developed some linguistic features like the number of characters, the difference between the number of characters, number of characters ratio etc.. They have used algorithms like XGBoost, Random Forest, AdaBoost and Decision Tree to develop the model for classification.

Indurthi et.al in their paper [18] state how they used linear regression on vector representation of posts created using word embeddings. They have prepared some hand-crafted features like presence of wh-words, checking if any digit is present at the starting of the title. Moreover, other notable features which they have used are: checking if any gerund or adjective in superlative form is present.

In the paper [13], Gairola et al. explains how they have created a predictive model by bringing together Bi-directional Long Short Term Memory Network (LSTM) with Attention and Siamese Neural Network on text and images. They have calculated character level embeddings using Convoluted Neural Networks (CNNs). They have combined this with distributed word embeddings and image embeddings achieving F1 score of 65.37%.

Philippe [28] have ignored the images and have used the only text by converting them to word embeddings. These word embeddings have been passed to a Long Short-Term Memory (LSTM) Network. This model's mean squared error is 0.0428, accuracy is 0.826, and an F1 score is 0.564.

Grigorev [16] also ignored images and used the Bag-of-Words method to convert text data to vectors. For capturing the sequential nature of the text data, he has further used bi-grams and tri-grams. For every text feature, a separate SVM regression model with a linear kernel is trained. These models are combined using Extremely Randomized Trees.

Glenski et. al [15] elaborates how LSTM and CNN have been applied separately after fusing text sequences with their 200 dimensional GloVe [25] representation. Zhou [29] have developed a multi-classification model. He applied self-attentive mechanism on hidden states of bi-directional GRUs. Bi-directional GRU have also been used by Omidvar et al. [23]. For designing the final layer they have used sigmoid activation to predict how much social media posts tend to be clickbait.

Ghosh in his paper [14] unfolds how classical machine learning techniques like Logistic Regression, Random Forests etc. and deep learning techniques like Deep Neural Networks, LSTM is used for detecting clickbaits. He has created several novel features like Word Mover's distances and cosine similarities between social media posts and actual news articles.

Morphological, Stylistic and Grammatical features have been used by Papadopoulou et al. [24]. They have created one model for each feature i.e. 65 models for 65 features in level 1. Outputs of these 65 models passed into a model in level 2 for classification. In the paper [27] , Rony et al. narrate how they have used distributed subword embeddings and topic modelling to develop a classifier capable of detecting click baits.

Presence of 2 kinds of Part-Of-Speech (POS) patterns, namely [number + noun phrase + verb] and [number +noun phrase + word "that"] is used as features by Cao et al. [5]. They have achieved AUC of 0.723 using Logistic Regression.

Anand et al. [1] re-defined the benchmarks set by state of the art models by achieving ROC-AUC of 0.99 using Recurrent Neural Networks (Bi directional LSTM, Bi directional GRU, Bi directional RNN) over Distributed Word Embeddings and Character Level Word Embeddings.

Biyani et al. [2] have developed various novel features like presence of superlative adverbs and adjectives, number of words, capital lettered words, presence of specific words (like "click here", "exclusive", "won't believe", "happens next", "don't want", "you know") and so on. Their F-1 score is 74.9%.

Chakraborty et al. in [7] elaborate on the process of generation and utilization of clickbait articles. In [6] they create features by analyzing structures of sentences(like lengths of the headlines, ratios of the number of stop words to the number of actual words and so on). They further inspect for patterns within words (like the presence of numbers, unexpected punctuations etc.). They also look for the presence of each word of the posts in a curated lexicon which comprises various clickbait phrases. Finally, they create a browser-based extension capable of detecting click bait which is 93% accurate.

Hu et al. in their paper [17] use convoluted neural networks for measuring similarities between 2 sentences.

Deudon presents the latest state of the artwork of measuring text similarity in NeurIPS 2018 [10]. He extends Word Mover's Distances proposed by Kusner et al. [21] using Variational Siamese Network.

# Chapter 3

# Understanding the data

## 3.1 Meta Data

Before preparing any model capable of detecting click baits, we need to understand the data first. This data has been obtained from Clickbait Challenge 2017 Workshop, Germany. It is prepared by Potthast et. al [26] by collecting tweets done by top news publishers.

### 3.1.1 Format & size of data

The raw data consists of 2 JSON files namely instances.jsonl, truth.jsonl and a folder media which contains images. Such files are present for each for training and validation. We merge them to form a single file for convenience. The instances.jsonl file contains information relating to social media posts and web pages they are referring to. The truth.jsonl file contains information about how much these posts tend to be clickbait. The size of the single merged CSV file without images is around 100 MB. The total size of the folder containing all images is around 1 GB.

### 3.1.2 Different attributes of data

This data consists of several independent and identically distributed instances. Each instance i.e. each row has several attributes / features / columns. They are described in detail in Table 3.1. The distribution of values of truthMean & truthMedian for click bait and non-clickbait posts are shown in 3.1. We see there is a huge overlap between truthMean values of clickbait and genuine posts. This is not so in the case of truthMedian.

| Variable Name | Variable Description |
|---|---|
| id | id of the instance / row |
| postMedia | path of the image files within the media folder |
| postText | text of the social media post |
| postTimestamp | date & time when the post was done |
| targetCaptions | Captions of images present in the actual web page |
| targetDescription | Description of the article in the actual web page |
| targetKeywords | Keywords from the article in the actual web page |
| targetParagraphs | Paragraphs of the article in the actual web page |
| targetTitle | Title of the article in the actual web page |
| truthClass | Whether the social media post is clickbait or not |
| truthJudgments | List of crowd sourced judgements relating to the post [not click baiting = 0.0, slightly click baiting = 0.33, considerably click baiting = 0.66, heavily click baiting = 1.0] |
| truthMean | Mean of the truthJudgements list |
| truthMedian | Median of the truthJudgements list |
| truthMode | Mode of the truthJudgements list |

Table 3.1 Variable Description

| # images | clickbait | no-clickbait | %-clickbait |
|---|---|---|---|
| 0 | 2827 | 7241 | 28.08% |
| 1 | 2601 | 8852 | 22.71% |
| 2 | 53 | 249 | 17.55% |
| 3 | 16 | 47 | 25.40% |
| 4 | 26 | 85 | 23.42% |

Table 3.2 Variation in % of click bait posts with # images

### 3.1.3 Exploratory Data Analysis

The data has 21,997 labeled instances. Around 16,474 i.e. 75% of these instances are not clickbaits, remaining 5523 instances i.e. 25% are clickbaits. The variation in percentage of click bait posts with number of images in them, week daywise, number of keywords in the target webpages and number of captions/images in the target webpages are shown is Tables 3.2, 3.3, 3.4, 3.5 respectively. We notice that the variation of percentage clickbait posts with days of the week the posts were done is not significant.

### 3.1.4 Training and Test split

We use a total of 18705 valid texts and 12690 valid images. We divide the data into mainly 2 parts. The first part is used for training a model capable of predicting the tendency post to be click bait. This is known as the training set. The second part is used to measure the accuracy of the model created. This part is called the test set. For developing text classifier and image classifier separately, we use around 67% of the data for training and the remaining 33% of the data for testing.

| weekday | clickbait | no-clickbait | %-clickbait |
|---------|-----------|--------------|-------------|
| Monday | 743 | 2283 | 24.55% |
| Tuesday | 712 | 2302 | 23.62% |
| Wednesday | 699 | 2360 | 22.85% |
| Thursday | 830 | 2536 | 24.66% |
| Friday | 848 | 2494 | 25.37% |
| Saturday | 896 | 2280 | 28.21% |
| Sunday | 795 | 2219 | 26.38% |

Table 3.3 Variation in % of click bait posts day wise

| # targetKeyWords | clickbait | no-clickbait | %clickbait |
|------------------|-----------|--------------|------------|
| 0 | 2124 | 5912 | 26.43% |
| 1 | 127 | 290 | 30.46% |
| 2 | 335 | 937 | 26.34% |
| 3 | 229 | 833 | 21.56% |
| 4 | 387 | 1151 | 25.16% |
| 5 | 452 | 1296 | 25.86% |
| 6 | 396 | 1365 | 22.49% |
| 7 | 353 | 1119 | 23.98% |
| 8 | 262 | 797 | 24.74% |
| 9 | 185 | 528 | 25.95% |
| 10 | 171 | 430 | 28.45% |

Table 3.4 Variation in % of click bait posts with number of target keywords upto 10

| # targetCaptions / targetImages | clickbait | no-clickbait | %clickbait |
|---------------------------------|-----------|--------------|------------|
| 0 | 868 | 2090 | 29.34% |
| 1 | 1168 | 4849 | 19.41% |
| 2 | 564 | 1937 | 22.55% |
| 3 | 427 | 1408 | 23.27% |
| 4 | 285 | 770 | 27.01% |
| 5 | 182 | 564 | 24.40% |
| 6 | 179 | 703 | 20.29% |
| 7 | 108 | 326 | 24.88% |
| 8 | 89 | 431 | 17.12% |
| 9 | 62 | 191 | 24.51% |
| 10 | 101 | 279 | 26.58% |

Table 3.5 Variation in % of click bait posts with number of target images upto 10

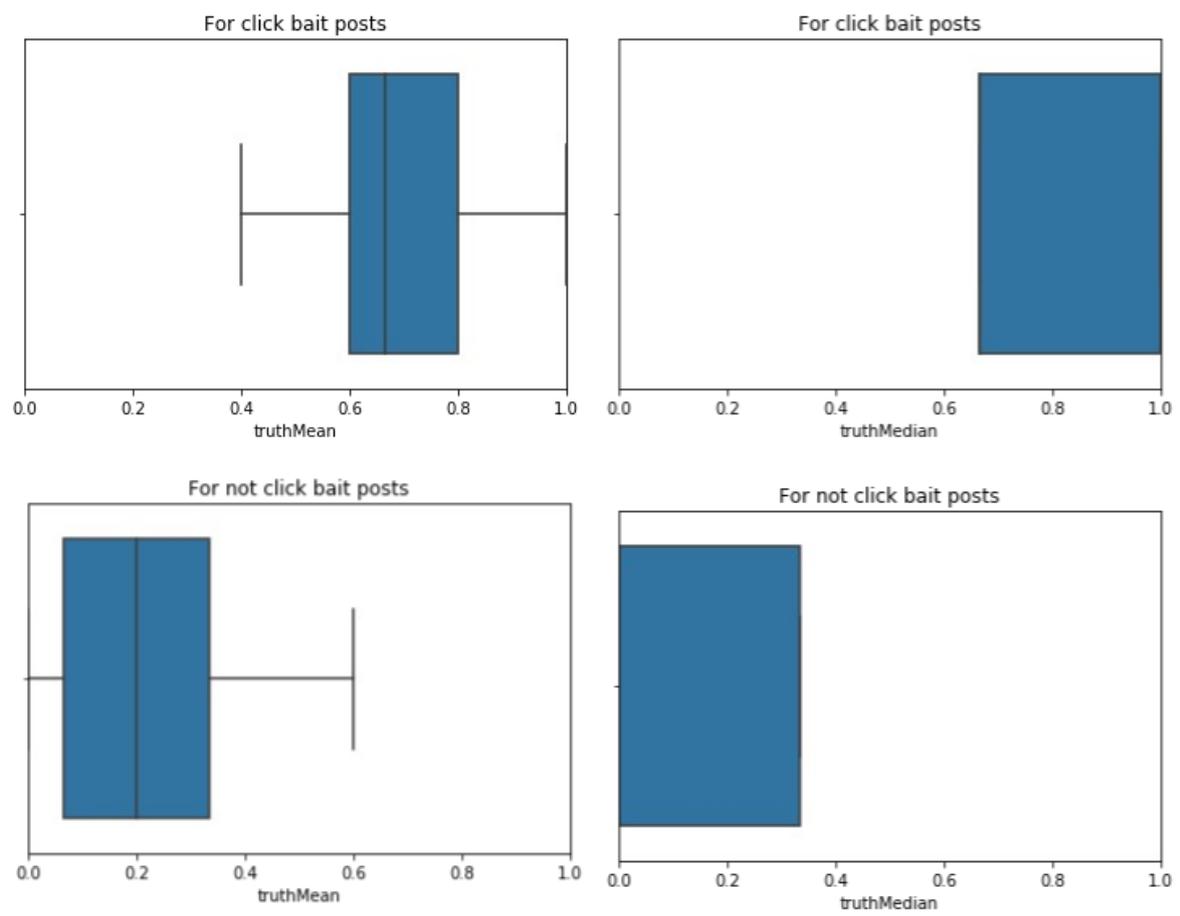

Fig. 3.1 Distribution of truthMean & truthMedian values for clickbait and non-clickbait articles

# Chapter 4

# Methodologies

Since Machine Learning models take only numbers as input, we need to extract numeric features from the texts and images. For developing Machine Learning based models we use a python library called sci-kit learn [4]. Another library named Keras [9] is used for building Deep Learning based models.

## 4.1   Feature Extraction from Texts

For extracting features from texts we transform each token / word of all text columns (like postText, targetCaptions, targetDescription,targetKeywords, targetParagraphs and targetTitle) into vectors.  Two kinds of algorithms are used for this :   Word2Vec[22] and GloVe[25]. However, the features extracted using word2vec of 50 dimensions have been dropped at a later stage since the model trained with only GloVe vectors of 50 dimensions exhibited better performance. Furthermore, for representing a sentence as a 50-dimensional vector, the median of GloVe vectors of tokens comprising it is taken. Only for training a LSTM based Deep Learning Model we have trained a word2vec model with 300 dimensions. While using Logistic Regression for classification, features which are highly correlated or having more than 90% entries as zeros are removed.

## 4.2   Engineering Features extracted from Texts

Feature engineering refers to the art of creating new features from existing features.  New features are generally introduced to increase the efficiency of the machine learning model. The new features that have been extracted from existing features of text data are mentioned in this section:

### 4.2.1  Creating Novel Features from Texts

- Word mover's distance [21] between (postText,targetTitle), (postText, targetDescription), (postText, targetParagraphs), (postText, targetKeywords), (postText, targetCaptions), (targetTitle, targetDescription), (targetTitle, targetParagraphs), (targetTitle, targetKeywords) and (targetTitle, targetCaptions)

- Polarity of the post extracted using TextBlob

- Polarity of targetCaptions extracted using TextBlob

- Polarity of targetDescription extracted using TextBlob

- Polarity of targetParagraphs extracted using TextBlob

- Polarity of targetTitle extracted using TextBlob

- Subjectivity of targetCaptions extracted using TextBlob

- Subjectivity of targetDescription extracted using TextBlob

- Subjectivity of targetParagraphs extracted using TextBlob

- Subjectivity of targetTitle extracted using TextBlob

- Cosine similarity between vector representations of postText & targetTitle

- Cosine similarity between vector representations of postText & targetDescription

- Cosine similarity between vector representations of postText & targetParagraphs

- Cosine similarity between vector representations of postText & targetKeywords

- Cosine similarity between vector representations of postText & targetCaptions

- Cosine similarity between vector representations of targetDescription & targetTitle

- Cosine similarity between vector representations of targetParagraph & targetTitle

- Cosine similarity between vector representations of targetKeywords & targetTitle

- Cosine similarity between vector representations of targetCaptions & targetTitle

### 4.2.2   Creating other Features

In addition to the novel features mentioned above, some other features have been extracted from this data following the works of various researchers. These features include:

- Number of captions of images in the target webpage

- Number of paragraphs in the target webpage

- Number of stop words in the post

- Number of unique punctuations in the post

- Count of tokens for each parts of speech in the postText

- Number of images in the postText

- Whether postText has any digits [18]

- Whether postText has any wh words in it[18]

- Whether any alluring words like "click here", "exclusive", "won't believe", "happens next", "don't want", "you know" is present [2]

- Jaccard similarity coefficient between words of posText and targetTitle

- Jaccard similarity coefficient between words of postText and targetDescription

## 4.3   Machine Learning based models for text classification

The label of this dataset i.e. truthClass is encoded as 0 if the post is not clickbait, else it is encoded as 1. We develop 4 classification models with different sets of parameters using all the features mentioned in sections 4.1 and 4.5. These models are based on the following algorithms:

- Logistic Regression

- Random Forest [3]

- XG-Boost [8]

- Light GBM [19]

### 4.3.1    Details of Logistic Regression Model

For the Logistic Regression model the following hyper-parameters are used:
**Random state:** 1,
**Class weight:** balanced

### 4.3.2    Details of Random Forest Model

For the Random Forest model the following hyper-parameters are used:
**Number of trees:** 200,
**Maximum depth:** 7
**Maximum features being considered at a time:**  19 (i.e.  square root number of features available)
**Random state:** 1

### 4.3.3    Details of XG-Boost Model

For the XG-Boost model the following hyper-parameters are used:
**Number of trees:** 5
**Learning rate:** 0.01
**Maximum depth:** 3
**Subsample rate:** 0.6
**Maximum percentage of features to be used in a tree:** 60%
**Regularized alpha:** 10

### 4.3.4    Details of Light GBM Model

For the Light GBM Model the following hyper-parameters are used:
**Number of trees:** 25
**Learning rate:** 0.01
**Maximum depth:** 7
**Subsample rate:** 0.2
**Maximum percentage of features to be used in a tree:** 30%
**Regularized alpha:** 5

## 4.4   Deep Learning based models for text classification

In this section,  various deep learning based models which have been used for detecting clickbait posts is discussed.  The label of this dataset i.e.  truthClass is encoded as 0 if the post is not clickbait, else it is encoded as 1.

### 4.4.1   Using Deep Neural Networks

For this Deep Neural Network based model all the features mentioned in sections 4.1 and 4.5 have been used. Figure 4.1 and represents this model.

Its' hyper-parameters are as follows:

**Number of sequential layers:** 3

**Details of Layer 1:** Number of Neurons = 50, Dropout = 0.2

**Details of Layer 2:** Number of Neurons = 300, Dropout = 0.3

**Details of Layer 3:** Number of Neurons = 2, Activation = softmax

**Loss:** Categorical Crossentropy

**Optimizer:** Adam [20]

```
Layer (type)                   Output Shape         Param #
=================================================================
dense_18 (Dense)               (None, 50)           19200
_________________________________________________________________
dropout_13 (Dropout)           (None, 50)           0
_________________________________________________________________
batch_normalization_5 (Batch   (None, 50)           200
_________________________________________________________________
dense_19 (Dense)               (None, 300)          15300
_________________________________________________________________
dropout_14 (Dropout)           (None, 300)          0
_________________________________________________________________
batch_normalization_6 (Batch   (None, 300)          1200
_________________________________________________________________
dense_20 (Dense)               (None, 2)            602
_________________________________________________________________
activation_6 (Activation)      (None, 2)            0
=================================================================
Total params: 36,502
Trainable params: 35,802
Non-trainable params: 700
_________________________________________________________________
```

Fig. 4.1 Architecture of Deep Neural Network based model

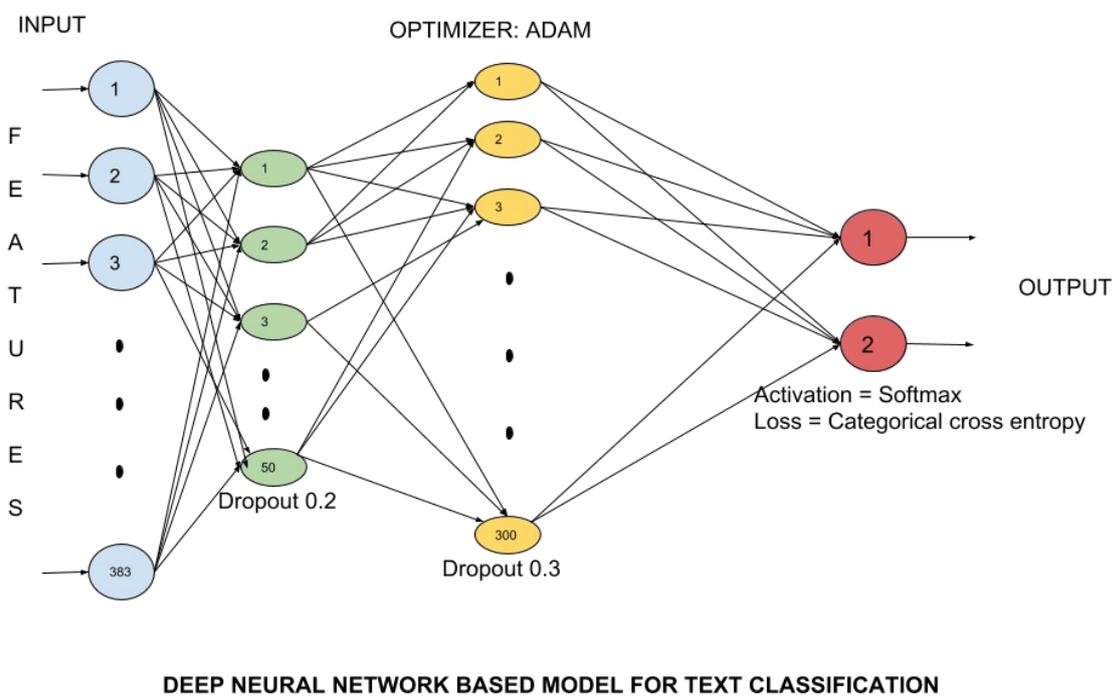

**DEEP NEURAL NETWORK BASED MODEL FOR TEXT CLASSIFICATION**

Fig. 4.2 Deep Neural Network based model for Text Classification

## 4.4.2   Using LSTM based Model

For this model, we create an embedding matrix using word2vec having 300 dimensions. Encoded and padded sequences of texts (postText, targetTitle, targetKeywords, targetDescription, targetParagraphs) have been generated using this embedding matrix. Each sequence has a maximum length of 140. Figure 4.3 represents this model. The architecture of the model is as follows:

**Number of sequential layers:** 4

**Details of Layer 1:** Layer Type: Embedding, weights are that of the embedding matrix

**Details of Layer 2:** Layer Type: LSTM, Units = 300, Dropout = 0.3, Recurrent dropout=0.3

**Details of Layer 3:** Number of Neurons = 400, Activation = ReLU, Dropout = 0.8

**Details of Layer 4:** Number of Neurons = 2, Activation = Softmax

**Loss:** Categorical Crossentropy

**Optimizer:** Adam [20]

**Early Stopping:** Monitor = Validation loss, Minimum delta=0, Patience=3, Verbose=0, Mode = auto

```
Layer (type)                 Output Shape              Param #
=================================================================
embedding_2 (Embedding)      (None, 700, 300)          60447000
_________________________________________________________________
spatial_dropout1d_2 (Spatial (None, 700, 300)          0
_________________________________________________________________
lstm_2 (LSTM)                (None, 300)               721200
_________________________________________________________________
dense_4 (Dense)              (None, 400)               120400
_________________________________________________________________
dropout_3 (Dropout)          (None, 400)               0
_________________________________________________________________
dense_5 (Dense)              (None, 2)                 802
_________________________________________________________________
activation_2 (Activation)    (None, 2)                 0
=================================================================
Total params: 61,289,402
Trainable params: 842,402
Non-trainable params: 60,447,000
_________________________________________________________________
```

Fig. 4.3 Architecture of LSTM based model

## 4.5   Feature Extraction from Images

To extract features from images, we first convert them into grey scale form using open cv library. Then, we convert them into arrays having 128 rows and 128 columns each.

## 4.6   Deep Learning based model for image classification

Figure 4.4 represents the CNN based deep learning model used for Image classification. The architecture of this model is as follows:

**Number of sequential layers:** 2

**Details of Layer 1:** Convoluted 2D Layer, Number of output filters = 10, Length of the 1D convolution window is 4X4, Pool size is 2 X 2

**Details of Layer 2:** Number of neurons = 1, Activation = Sigmoid

**Loss:** Binary Crossentropy

**Optimizer:** Adam [20]

**Early Stopping:** Monitor = Validation loss, Minimum delta=0, Patience=3, Verbose=0, Mode = auto

```
Layer (type)                    Output Shape              Param #
=================================================================
conv2d_24 (Conv2D)              (None, 125, 125, 10)      170

max_pooling2d_24 (MaxPooling    (None, 62, 62, 10)        0

flatten_18 (Flatten)            (None, 38440)             0

dense_32 (Dense)                (None, 1)                 38441
=================================================================
Total params: 38,611
Trainable params: 38,611
Non-trainable params: 0
```

Fig. 4.4 Architecture of CNN based model for image classification

# Chapter 5

# Results

In this section, we present the performances of the models discussed in the previous chapter. We use metrics like Accuracy, Mean Squared Error (MSE), Area under ROC curve, Precision, Recall and F1-Score. The results of Text Classification models have been presented in Table 5.1 and that of Image Classification models have been presented in Table 5.2. Furthermore, the ROC curves for Logistic Regression, Random Forests, XG-Boost & Light GBM models have been depicted in Figures 5.1, 5.2, 5.3 and 5.4 respectively. Figures 5.5 and 5.6 are the ROC curves for Deep Neural Network based and LSTM based Text Classification Models. The image Classification Model's performance is shown by the ROC curve 5.10. Moreover, the top 30 variables in terms of importance have been depicted in Figures 5.7, 5.8 and 5.9 for Random Forest, XG-Boost and Light GBM text classification models respectively. Each of these results will be discussed in details in the next chapter.

| Text Classification Model | Dataset | Accuracy | MSE | AUC | Precision | Recall | F1-Score |
|---|---|---|---|---|---|---|---|
| Logistic Regression | Train | 0.79 | 0.21 | 0.87 | 0.83 | 0.79 | 0.8 |
| Logistic Regression | Test | 0.78 | 0.22 | 0.85 | 0.82 | 0.78 | 0.79 |
| Random Forest | Train | 0.81 | 0.17 | 0.87 | 0.84 | 0.83 | 0.8 |
| Random Forest | Test | 0.81 | 0.19 | 0.83 | 0.81 | 0.81 | 0.77 |
| XG-Boost | Train | 0.82 | 0.18 | 0.83 | 0.82 | 0.82 | 0.79 |
| XG-Boost | Test | 0.81 | 0.19 | 0.81 | 0.8 | 0.81 | 0.77 |
| Light GBM | Train | 0.76 | 0.24 | 0.87 | 0.58 | 0.76 | 0.66 |
| Light GBM | Test | 0.76 | 0.24 | 0.83 | 0.58 | 0.76 | 0.66 |
| Deep Learning - DNN | Train | 0.89 | 0.08 | 0.94 | 0.89 | 0.89 | 0.89 |
| Deep Learning - DNN | Test | 0.82 | 0.12 | 0.85 | 0.82 | 0.82 | 0.82 |
| Deep Learning - LSTM | Train | 0.78 | 0.17 | 0.66 | 0.76 | 0.78 | 0.73 |
| Deep Learning - LSTM | Test | 0.78 | 0.17 | 0.64 | 0.76 | 0.78 | 0.73 |

Table 5.1 Performance of different Text Classification Models

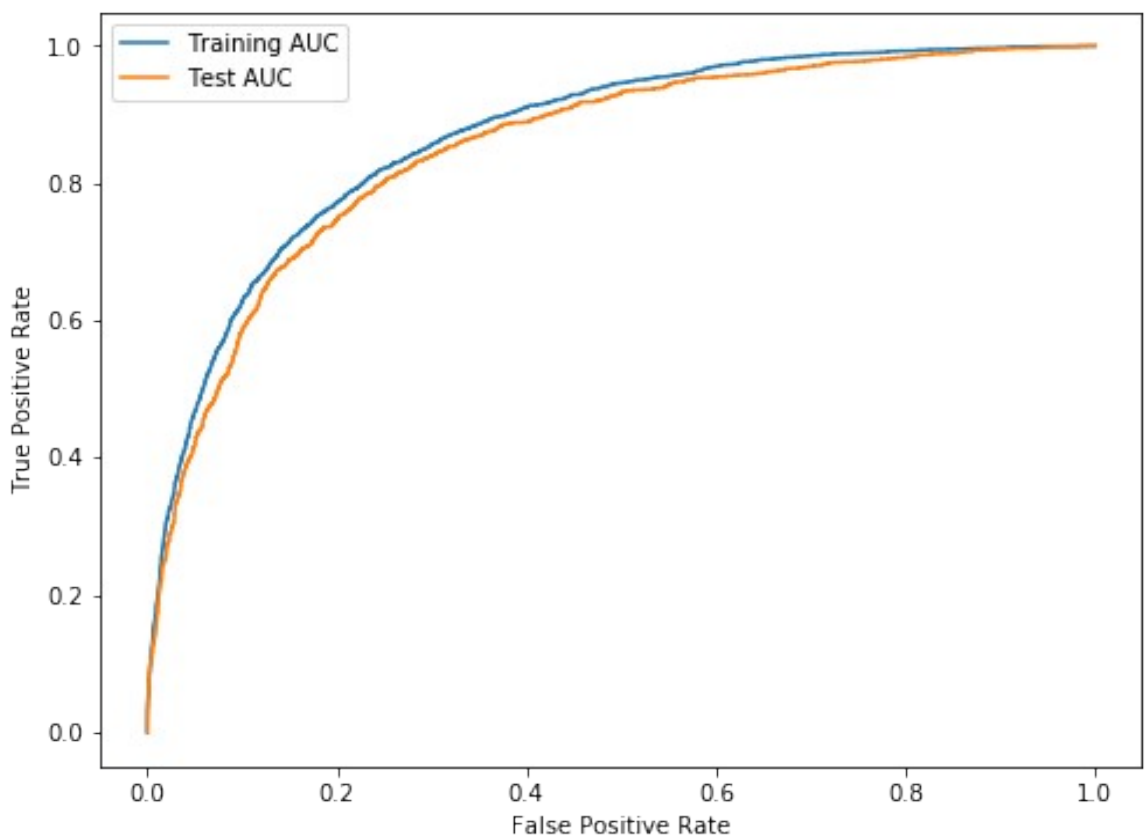

Fig. 5.1 ROC curve for Logistic Regression (Text)

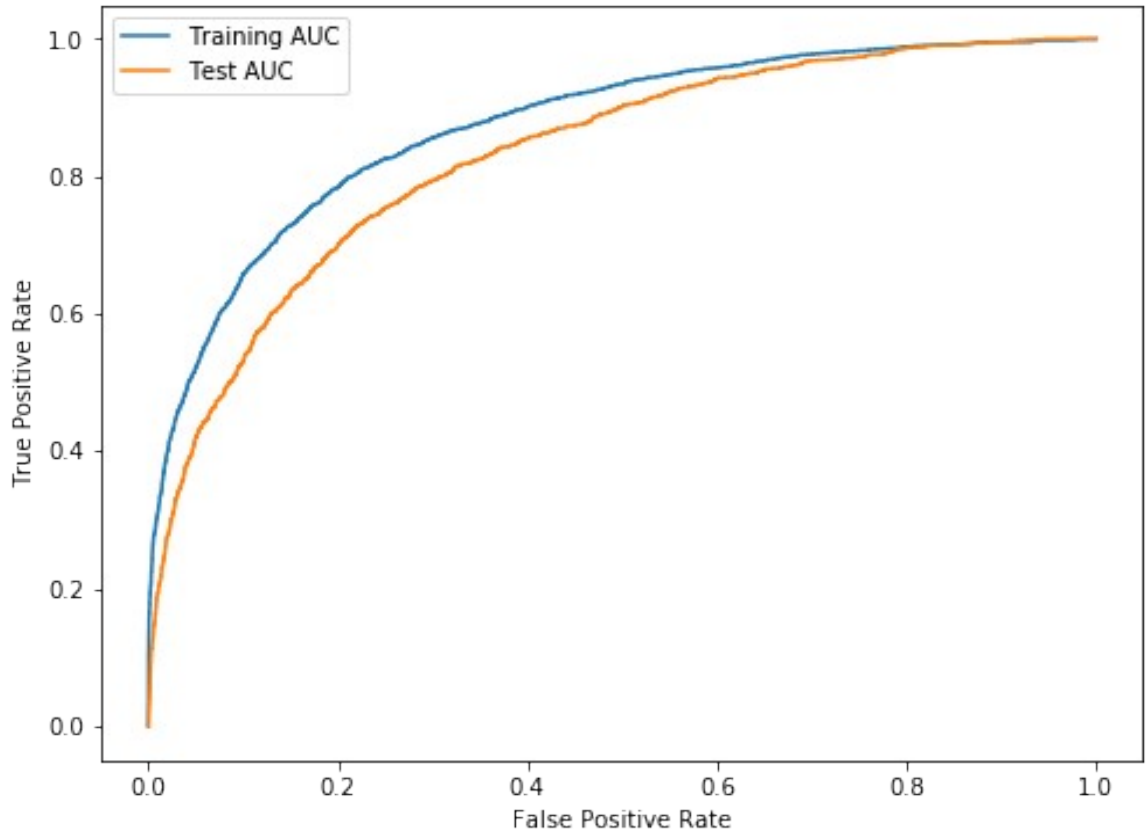

Fig. 5.2 ROC curve for Random Forest (Text)

| Image Classification Model | Dataset | Accuracy | MSE | AUC | Precision | Recall | F1-Score |
|---|---|---|---|---|---|---|---|
| **Deep Learning** | Train | 0.83 | 0.12 | 0.87 | 0.84 | 0.83 | 0.79 |
| **Deep Learning** | Test | 0.77 | 0.18 | 0.72 | 0.6 | 0.77 | 0.68 |

Table 5.2 Performance of Deep Learning based Image Classification Model

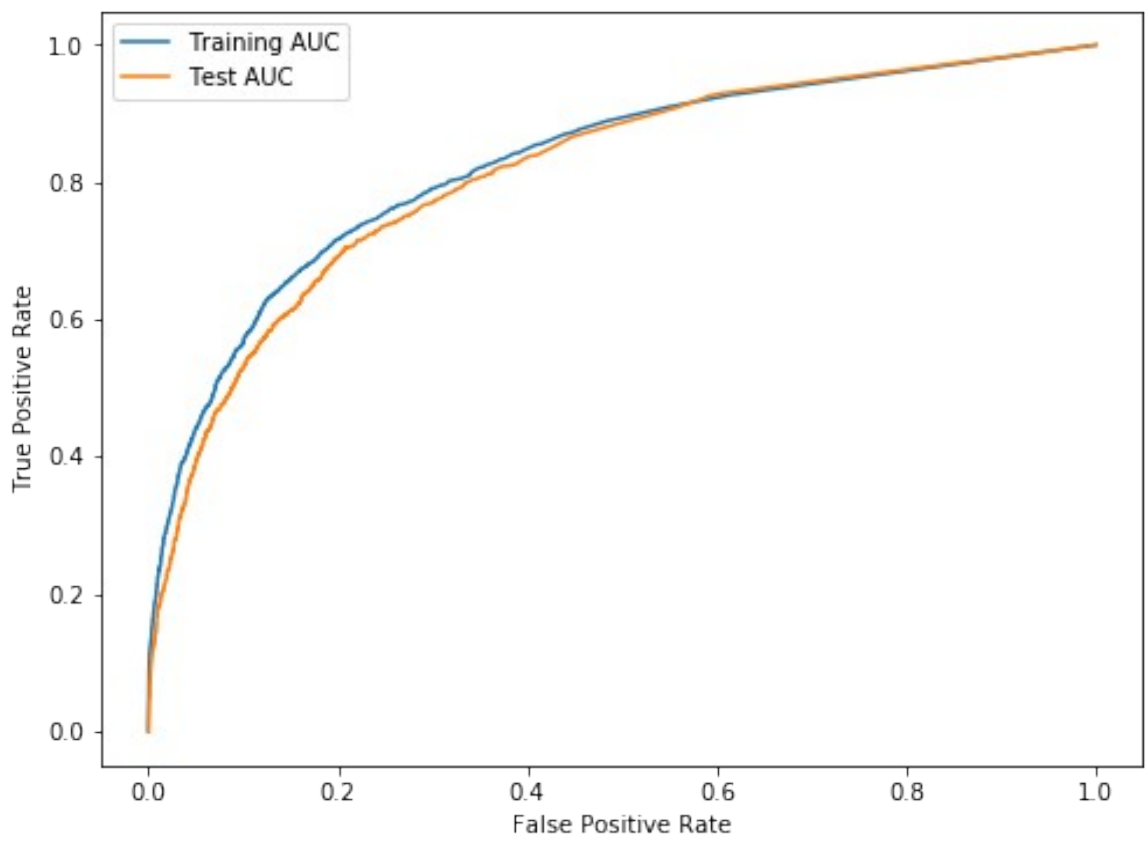

Fig. 5.3 ROC curve for XG-Boost (Text)

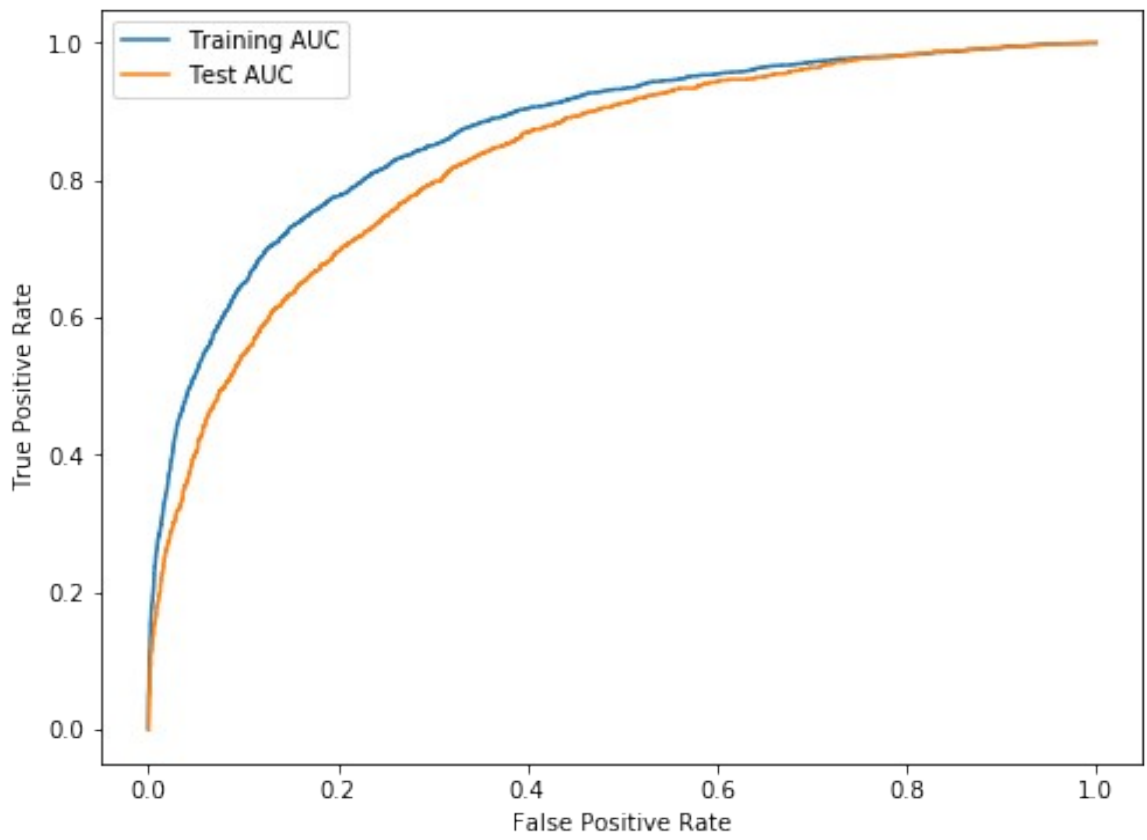

Fig. 5.4 ROC curve for Light-GBM (Text)

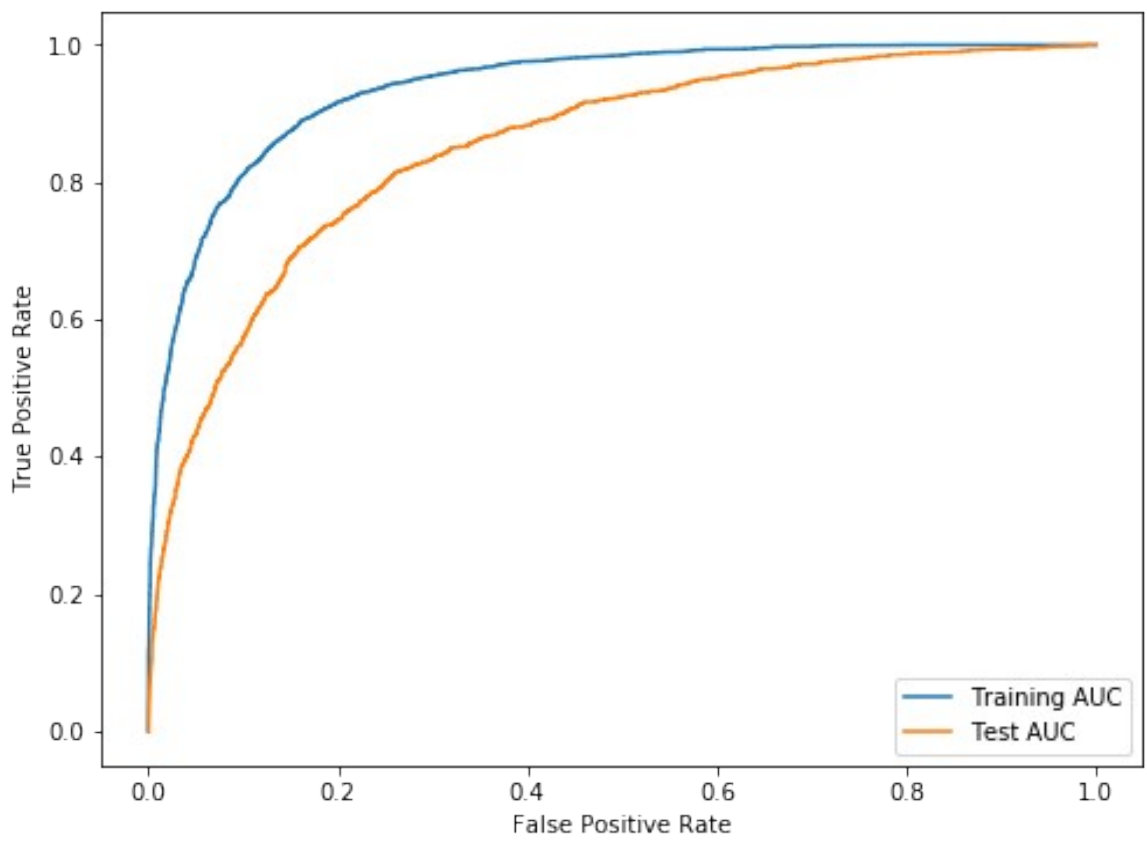

Fig. 5.5 ROC curve for Deep Neural Network based Model (Text)

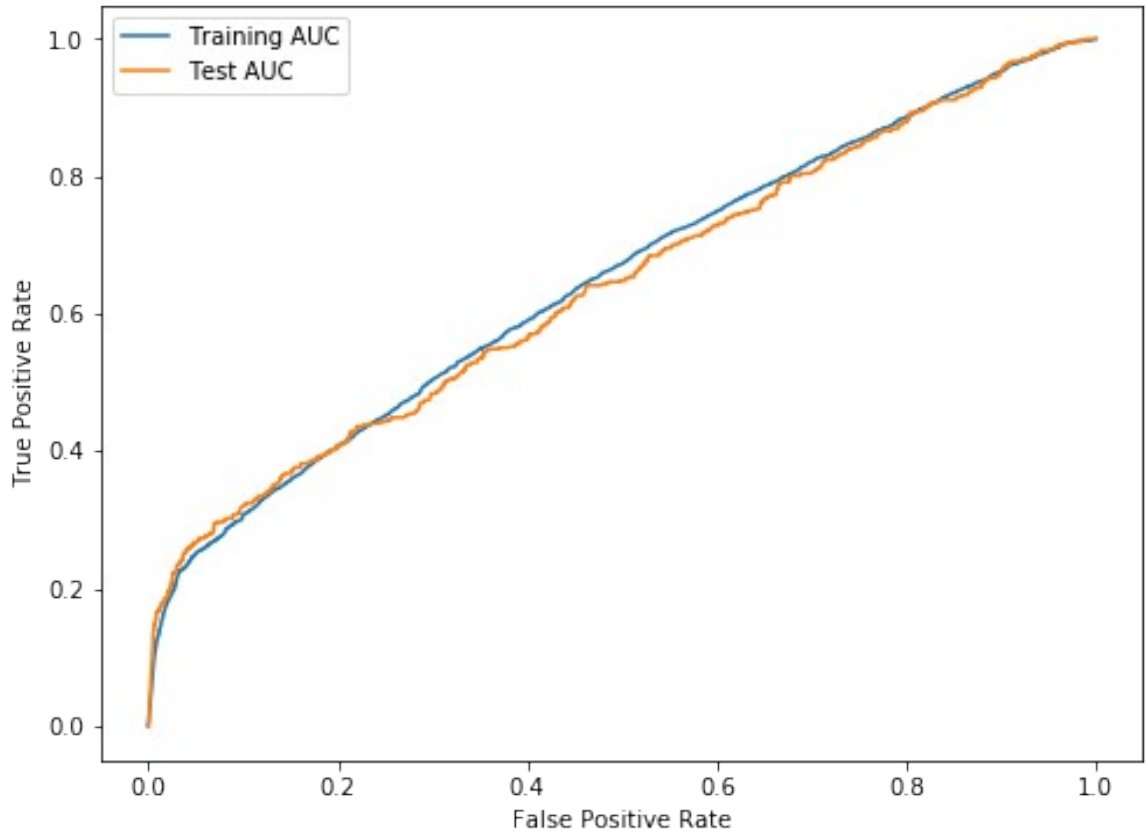

Fig. 5.6 ROC curve for LSTM based Model (Text)

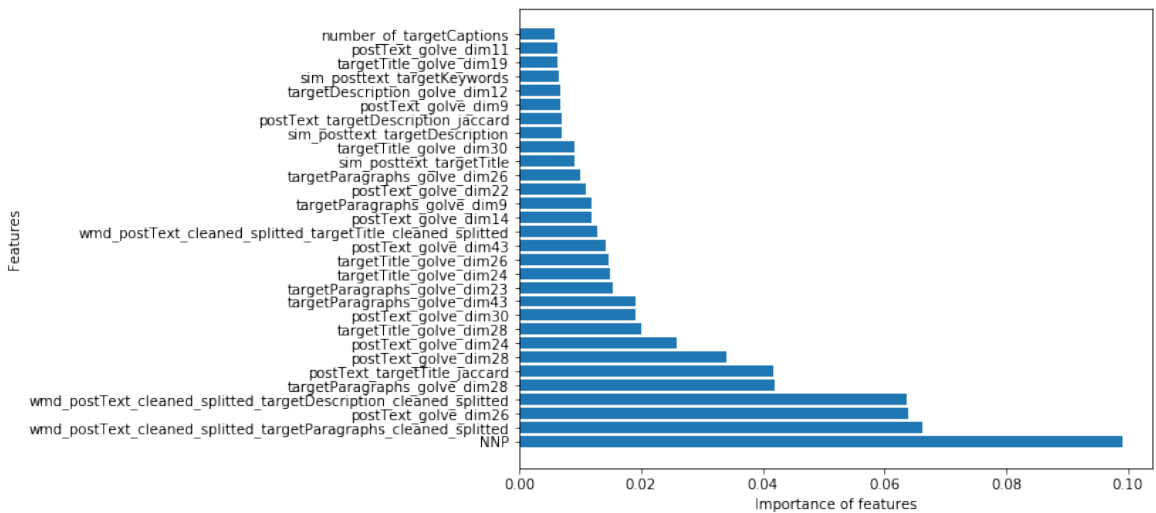

Fig. 5.7 Variable Importances from Random Forest

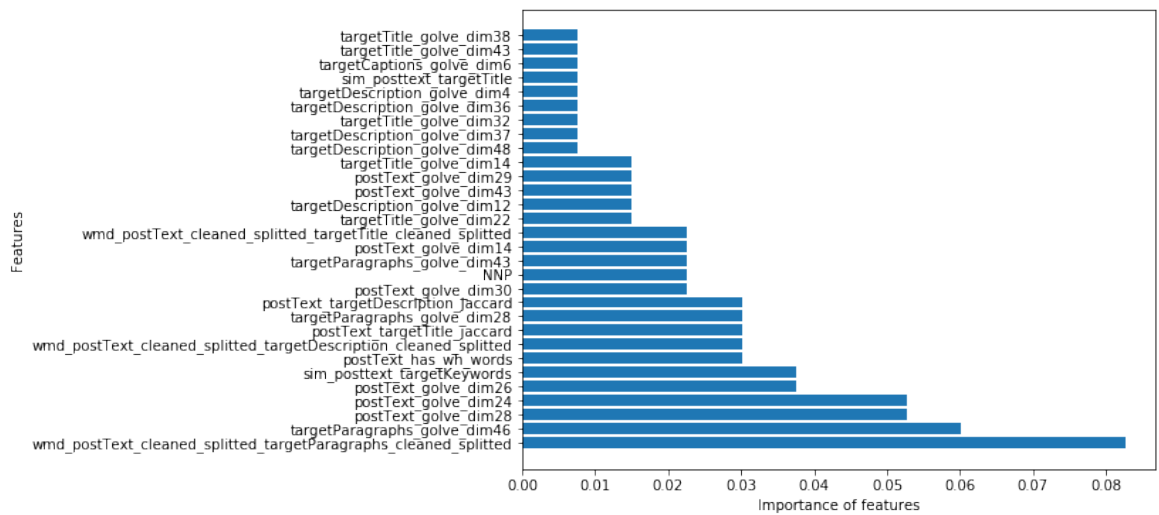

Fig. 5.8 Variable Importances from XG-Boost

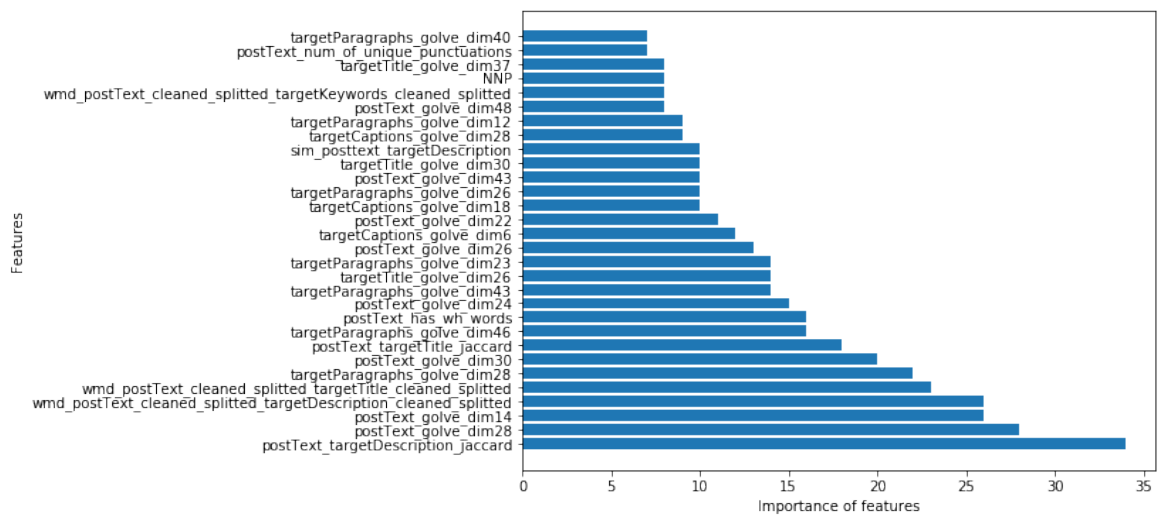

Fig. 5.9 Variable Importances from Light GBM

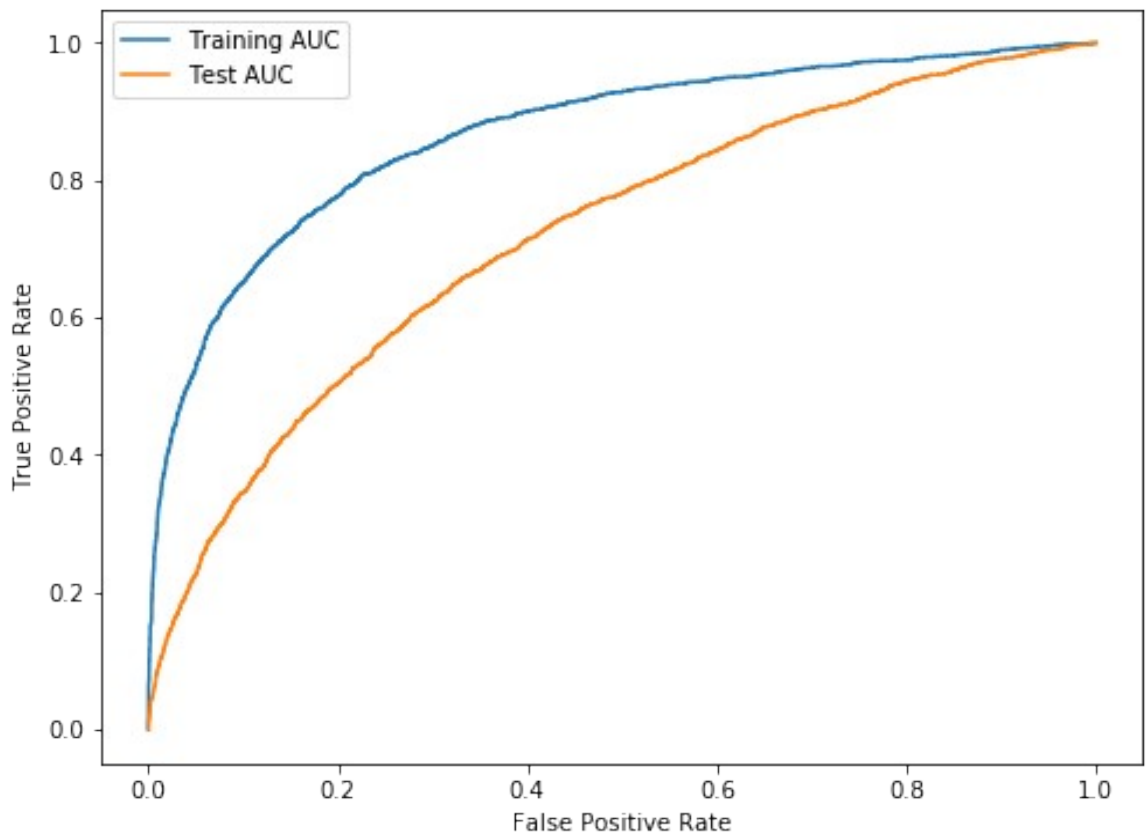

Fig. 5.10 ROC curve for Image Classification Model

# Chapter 6

# Discussions and Conclusions

## 6.1    Performances of Models

From Table 5.1 it can be inferred that for Text Classification, Deep Neural Network based model results in the maximum area under the ROC curve, least MSE and highest F1-score. But, observing the difference between performances of this model in the training and test set, we can conclude that it over-fits a bit. Furthermore, this model is very complex as it consists of 35,802 trainable parameters. If we deploy want to this model in production we have to compromise with explainability. This is not acceptable by the product (business) team. Thus, we choose the second best model i.e. Logistic Regression model. It is simple and can be easily explained to the business team. Although its' MSE is somewhat higher than the previous model, their areas under ROC curves are comparable. Furthermore, unlike the former one, this Logistic Regression based model does not over-fit. Hence, we choose this model for deployment in production.

On comparing top 5 important variables using plots 5.7, 5.8 and 5.9 it is interesting to note that not a single variable is common among them. However, when we consider the top 10 important variables of each model we find four common variables. They are the 28th dimension of GloVe representation of postText and targetParagraphs, Jaccard similarity coefficient between postText and targetTitle, Word Mover's Distance between postText and targetDescription. Furthermore, we observe that among these top 10 important variables, some variables like presence of proper noun in singular form, 28th dimension of GloVe representation of targetTitle are exclusively appears for the RF model. Similarly, cosine similarity between postText and targetKeywords is exclusive for XGB model. For the LGBM model, the 14th dimension of GloVe representation of postText, Jaccard similarity coefficient between postText and targetDescription and Word Mover's Distance between postText and targetTitle are exclusive. These variations arise due to the difference in the architecture of these models.

There are various other Deep Learning based models which have been tried for classifying texts. They include Bi-directional LSTM, GRUs, Bi-directional LSTM based Siamese Network with Arrention and different other variations of RNNs. Since they are not performing up to the mark, we are not mentioning them here.

The Image Classification model's performance is average. We shall definitely try to improve this in the next iteration. Various other classical machine learning based approaches (like LR, RF, XGB, LGBM) and deep learning based approaches (like different variations of CNNs, transfer learning using VGG19) have been tried. We are only showcasing the most efficient model here.

## 6.2 Benchmarking performances of Models

Comparing our results mentioned in Table 5.1 with those present in Clickbait Challenge 2017 we observe our deep neural network based text classification model performs better than 2 out of top 16 models in terms of MSE, 5 out of top 16 models in terms of Accuracy, 16 out of top 16 models in terms of F1 score and Precision, 14 out of top 16 models in terms of Recall. Moreover, the Logistic Regression model we are deploying performs better than 1 out of top 16 models in terms of MSE, 4 out of top 16 models in terms of Accuracy, 16 out of top 16 models in terms of F1 score and Precision, 14 out of top 16 models in terms of Recall.

## 6.3 Deploying Models in production

After discussion with the business team, it has been finalized to deploy the Logistic Regression based Text Classifier Model in production. The simplicity of this model is worth noticing. Presently, we are deploying only the text classifier model as the image classifier's performance is mediocre. In the next iteration, we shall deploy the image classifier and ensemble its' results with the text classifier. This prototype is deployed using python. At this moment it is taking around 7.1 minutes to generate a single prediction on a Intel(R) Xeon(R) CPU E5-2609 processor server with 181 GB RAM. Out of this 7.1 minutes, the vector representation of texts using GloVe takes around 7 minutes. Other steps take less than 1 second. In future, we shall surely ensure that the system can respond in real time. For this, incorporating multi-threading may be essential. For enhancing its' performance further, we shall keep the Logistic Regression model and GloVe vectors loaded in memory. The input form like Fig. 6.1.

Enter number of pictures/captions in the target (i.e. news article):

Enter number of paragraphs in the target (i.e. news article):

Please enter post i.e tweet:

enter text here

Please enter title of the news article:

enter text here

Please enter short description of the news article:

enter text here

Please enter paragraphs of the news article:

enter text here

Please enter keywords/tags relating to the news artices separated by commas:

enter text here

Please enter captions of images present in the news article separated by commas:

enter text here

Submit

<div align="center">Fig. 6.1 Input Form</div>

## 6.4   Future Work

Although the present state of the model is acceptable, still we want to improve it further. To do so, we shall be trying the following:

- Using tweet tokenizer to parse tweets. It is expected to extract tokens/words from posts more efficiently.

- Presently, stop words have not been removed while representing sentences in vector space. This has been done to keep the sense of a sentence intact. However, removing stop words may increase the overall efficiency of the model.

- As of now, Glove word embeddings with 50 dimensions and word2vec embeddings with 300 dimensions have been used. We shall try more dimensions. Increasing the number of dimensions will help us to represent sentences effectively. But, at the same time, the performance of the model needs to be compromised.

- From, Table 5.2 we observe that our Deep Learning based Image classification model is not state of the art. We shall try using Transfer Learning and Generative Adversarial Networks to improve its' accuracy.

- Recently, transfer learning has also become immensely popular in the field of NLP. Using pre-trained models (like BERT- Bidirectional Encoder Representations) [11] along with word vectors may yield better results.

- Presently, we have used various similarity measures (like word mover's distance [21], cosine similarity and so on) to calculate similarities between texts. In future, we will use a recently proposed state of the art similarity measure which uses Variational Siamese Network [10].

**Checklist of items for the Final Dissertation Report**
This checklist is to be attached as the last page of the report.

This checklist is to be duly completed, verified and signed by the student.

| | | |
|---|---|---|
| 1. | **Is the final report neatly formatted with all the elements required for a technical Report?** | Yes / No |
| 2. | Is the Cover page in proper format as given in Annexure A? | Yes / No |
| 3. | Is the Title page (Inner cover page) in proper format? | Yes / No |
| 4. | (a) Is the Certificate from the Supervisor in proper format? | Yes / No |
| | (b) Has it been signed by the Supervisor? | Yes / No |
| 5. | Is the Abstract included in the report properly written within one page? Have the technical keywords been specified properly? | Yes / No <br> Yes / No |
| 6. | Is the title of your report appropriate? **The title should be adequately descriptive, precise and must reflect scope of the actual work done.** Uncommon abbreviations / Acronyms should not be used in the title | Yes / No |
| 7. | Have you included the List of abbreviations / Acronyms? | Yes / No |
| 8. | Does the Report contain a summary of the literature survey? | Yes / No |
| 9. | Does the Table of Contents include page numbers? | Yes / No |
| | (i).   Are the Pages numbered properly? (Ch. 1 should start on Page # 1) | Yes / No |
| | (ii).  Are the Figures numbered properly? (Figure Numbers and Figure Titles should be at the bottom of the figures) | Yes / No |
| | (iii). Are the Tables numbered properly? (Table Numbers and Table Titles should be at the top of the tables) | Yes / No |
| | (iv).  Are the Captions for the Figures and Tables proper? | Yes / No |
| | (v).   Are the Appendices numbered properly? Are their titles appropriate | Yes / No |
| 10. | Is the conclusion of the Report based on discussion of the work? | Yes / No |
| 11. | Are References or Bibliography given at the end of the Report? <br> Have the References been cited properly inside the text of the Report? <br> Are all the references cited in the body of the report | Yes / No <br> Yes / No <br> Yes / No |
| 12. | Is the report  format and content according to the guidelines? The report should not be a mere printout of a Power Point Presentation, or a user manual. Source code of software need not be included in the report. | Yes / No |

**Declaration by Student:**
I certify that I have properly verified all the items in this checklist and ensure that the report is in proper format as specified in the course handout.

*Sohom Ghosh*
**Signature of the Student**

**Place:**  Noida

**Date :** 28th March 2019

**Name:**  Sohom Ghosh

**ID No.:**  2017HT12194